\documentclass[journal,twoside]{IEEEtran}
\usepackage{cite}
\usepackage{amsmath,amssymb,amsfonts}
\usepackage{algorithm}
\usepackage{algorithmic}
\usepackage{float}
\usepackage{graphicx}
\usepackage{textcomp}
\usepackage{wrapfig}
\usepackage{multirow}
\usepackage{booktabs}
\usepackage{makecell}
\usepackage{placeins}
\usepackage{hyperref}
\hypersetup{
            hidelinks=true, breaklinks=true, bookmarksnumbered=true, bookmarksopen=true}
\renewcommand{\skew}[1]{\left[#1\right]_{\times}}

\begin{document}
\title{Analytical Covariance Propagation for DVL-Aided Loosely Coupled SINS Under Attitude Uncertainty}
\author{Jin Huang, , Zichen Liu, Haoda Li, Zhikun Wang, Yuan Zhao, Xianyu Peng, Yongqun Yu, and Ying Chen
\thanks{Corresponding authors: Zhikun Wang and Ying Chen.}
\thanks{Jin Huang, Zichen Liu, Haoda Li, Yuan Zhao, Xianyu Peng, Yongqun Yu, and Ying Chen are with the State Key Laboratory of Ocean Sensing, Ocean College of Zhejiang University, Zhoushan, 316021, China (e-mail: jin.huang@zju.edu.cn; zichen\_liu@zju.edu.cn; haodleo@zju.edu.cn; ocean\_zhao@zju.edu.cn; 22434069@zju.edu.cn; 12534125@zju.edu.cn; ychen@zju.edu.cn).}
\thanks{Zhikun Wang is with the Donghai Laboratory, Zhoushan, 316021, China (e-mail: wang\_zk@zju.edu.cn).}
}

\IEEEaftertitletext{%
\begingroup
\setlength{\fboxrule}{0.4pt}%
\setlength{\fboxsep}{8pt}%
\noindent\fbox{%
\parbox{\dimexpr\textwidth-2\fboxsep-2\fboxrule\relax}{%
\textbf{IEEE Copyright Notice}\par
This is the accepted manuscript version of an article accepted for publication in \textit{IEEE Sensors Journal}. \copyright\ 2026 IEEE. Personal use of this material is permitted. Permission from IEEE must be obtained for all other uses, in any current or future media, including reprinting/republishing this material for advertising or promotional purposes, creating new collective works, for resale or redistribution to servers or lists, or reuse of any copyrighted component of this work in other works.}}
\endgroup
\vspace{1em}}

\maketitle
\begin{abstract}
In loosely coupled strapdown inertial navigation system/Doppler velocity log (SINS/DVL) integration, the body-frame DVL velocity is projected into the navigation frame using the SINS-computed attitude.
Attitude uncertainty consequently affects both the projected velocity observation and its associated measurement covariance.
This paper develops an analytical covariance propagation (ACP) method that treats these two effects consistently.
The attitude-error-aware observation model represents perturbations in the projected DVL velocity, whereas the covariance construction propagates the body-frame DVL covariance at the matrix level and adds a closed-form term derived from the predicted attitude-error covariance, rather than directly rotating component-wise standard deviations.
Simulation and a controlled surface-water field experiment compare ACP with conventional SINS/DVL and adaptive-covariance benchmarks.
For the evaluated data sets and parameterizations, ACP yields the lowest position-error metrics among the compared methods. Relative to the variational Bayesian adaptive Kalman filter (VBAKF) in the field experiment, ACP reduces the northward, eastward, and downward position root-mean-square errors (RMSEs) by 48.5\%, 55.0\%, and 74.8\%, respectively.
Beyond DVL aiding, ACP provides a transferable covariance-construction principle for attitude-dependent vector-measurement fusion and a physically interpretable alternative to heuristic frame-dependent covariance assignment.
\end{abstract}

\bstctlcite{IEEEtranBSTCTL}

\section{Introduction}

Autonomous underwater vehicles (AUVs) are increasingly used for ocean exploration, environmental monitoring, seabed mapping, and underwater infrastructure inspection \cite{paullAUVNavigationLocalization2014, terraccianoMarineRobotsUnderwater2020, ioannouUnderwaterInspectionMonitoring2024, liHybriddrivenDiscshapedAutonomous2026}.
Reliable navigation is essential for these missions, but global navigation satellite system (GNSS) signals are generally unavailable underwater because electromagnetic waves attenuate rapidly in water \cite{zhangAutonomousUnderwaterVehicle2023, maurelliAUVLocalisationReview2022, huangPreciseTimeDelay2026a}.
Underwater vehicles therefore rely on inertial navigation aided by complementary sensors, such as Doppler velocity logs (DVLs), acoustic positioning systems, and terrain- or map-aided information, depending on mission requirements and sensor availability \cite{millerAutonomousUnderwaterVehicle2010, engelsmanInformationAidedInertialNavigation2023, huangRaspi2USBLOpenSourceRaspberry2026}.

A strapdown inertial navigation system (SINS) provides continuous position, velocity, and attitude information without external infrastructure and is therefore a core component of many AUV navigation systems.
Because gyroscope and accelerometer measurements are integrated through the navigation mechanization, however, SINS errors accumulate with time \cite{grovesPrinciplesGNSSInertial2013, tittertonStrapdownInertialNavigation2004}.
DVL velocity aiding is widely used to constrain this drift because it provides relative velocity measurements with respect to the seabed or the water mass, and integrated-navigation experiments have demonstrated its practical value in underwater applications \cite{jalvingDVLVelocityAiding2004, kinseyPreliminaryFieldExperience2004}.
A DVL estimates velocity from acoustic Doppler measurements and can provide beam-level, instrument-frame, or body-frame velocity information \cite{whitcombAdvancesDopplerBased1999}.
In SINS/DVL integration, loosely coupled filters usually fuse a DVL-derived velocity vector with the SINS navigation solution, whereas tightly coupled filters directly use beam-level measurements and are more suitable for partial DVL observations \cite{talINSDVLFusion2017, wangNovelSINSDVL2020}.

Existing studies have improved SINS/DVL navigation from several complementary directions.
Methods for degraded DVL availability address DVL outages, limited beam returns, and data-driven beam enhancement \cite{kleinContinuousINSDVL2020, cohenBeamsNet2022, cohenSeamlessUnderwaterNavigation2024}.
Environmental and sensing-mode studies consider water-track operation, seafloor gradient, ocean current, and track-model mismatch \cite{hegrenaesDopplerWaterTrack2009, braginskyCorrectionDVLError2020, huangEstimationConstantOcean2023, yaoModifiedSmoothingScheme2023, wangNovelSINSDVL2024}.
Calibration-oriented methods estimate DVL scale factors, installation angles, lever arms, and sensor-alignment parameters through in situ, robust, online, and position-aided formulations \cite{kinseyInSituAlignment2007, xuNovelDVLCalibration2022, wangOnlineCalibrationMethod2022a, zhangCalibrationMethodDVL2024a, huangGNSSaidedInstallationError2025}.
DVL-aided alignment and attitude-error modeling studies address attitude-dynamics-induced velocity errors, robust in-motion or initial alignment, and body-frame attitude-error formulations for SINS/DVL integration \cite{liuCorrectionMethodDVL2017, yaoInMotionCoarseAlignment2020, xuRobustInMotionAlignment2021, hongyangRobustAdaptiveSINS2024, gaoEulerAngleError2025}.
These studies address measurement availability, environmental disturbance, sensor calibration, alignment, and deterministic velocity correction.
Although attitude errors have been incorporated into DVL-aided navigation models, they are mainly treated through deterministic observation equations, alignment procedures, or calibration models.
Less attention has been paid to the stochastic consistency of projecting both a body-frame DVL velocity and its covariance into the navigation frame when the SINS-computed attitude is uncertain.
In this setting, attitude uncertainty perturbs the projected velocity observation and changes the measurement covariance used to weight that observation.

Kalman filters assign measurement weights through the measurement-noise covariance matrix \cite{andersonOptimalFiltering2005, barShalomEstimationApplications2002}. Under a coordinate transformation, uncertainty should be propagated at the covariance-matrix level \cite{thrunProbabilisticRobotics2005, julierUnscentedFiltering2004}. However, representative loosely coupled SINS/DVL formulations often prescribe a diagonal or axis-wise covariance for the navigation-frame velocity update \cite{liNovelSchemeDVLAided2013}. Other approaches adapt the covariance online from innovation or residual statistics, including recursive noise estimation, Sage-Husa filtering, variational Bayesian filtering, and event-triggered schemes \cite{gaoAdaptiveKalmanFiltering2015, liuModifiedSageHusa2021, zhuRobustFilterSINSDVL2024, maEventTriggeredState2025}. In contrast, this paper addresses the pre-update construction of the projected DVL covariance from the body-frame noise model and the uncertainty in the attitude used for projection.

This paper investigates analytical covariance propagation (ACP) for DVL-aided loosely coupled SINS navigation under attitude uncertainty.
The objective is not to replace DVL outage handling, beam-level fusion, calibration, or adaptive filtering.
Instead, the proposed method addresses a complementary stochastic-consistency problem: how to keep the projected DVL velocity and its transformed measurement covariance consistent when the projection depends on an uncertain attitude.
ACP propagates a general, possibly non-diagonal, body-frame DVL noise covariance through the attitude-uncertain velocity transformation by using the predicted attitude-error covariance.

The principal contributions and evaluation scope are summarized as follows.
\begin{enumerate}
\item The projection of a DVL velocity measurement under attitude uncertainty is formulated as a stochastic-consistency problem in which the attitude error affects both the projected observation and its measurement covariance.
\item The proposed ACP method treats these effects consistently. The attitude-error-aware observation model accounts for perturbations in the projected velocity, and the covariance construction combines the nominal matrix-level transformation with a closed-form attitude-uncertainty term derived from the predicted attitude-error covariance. No additional residual-driven adaptation gain is introduced.
\item ACP is evaluated against conventional SINS/DVL and a SINS/DVL (Sage-Husa) adaptive-covariance benchmark in both simulation and a controlled surface-water field experiment. The field experiment additionally includes SINS/DVL (VBAKF) as a second adaptive-covariance benchmark.
\end{enumerate}

The remainder of this paper is organized as follows.
Section~\ref{secBriefReview} summarizes the loosely coupled SINS/DVL navigation model.
Section~\ref{secProposedMethod} presents the attitude-error-aware DVL observation formulation and the analytical covariance propagation method.
Section~\ref{secResultDiscussion} reports the simulation results, the field-experiment results, and the comparative analysis.
Finally, Section~\ref{secConclusion} concludes this paper.

\section{Loosely Coupled SINS/DVL Navigation Model}
\label{secBriefReview}
\subsection{System Structure and Error-State Model}

\begin{figure}
  \centering
  \includegraphics[width=\linewidth]{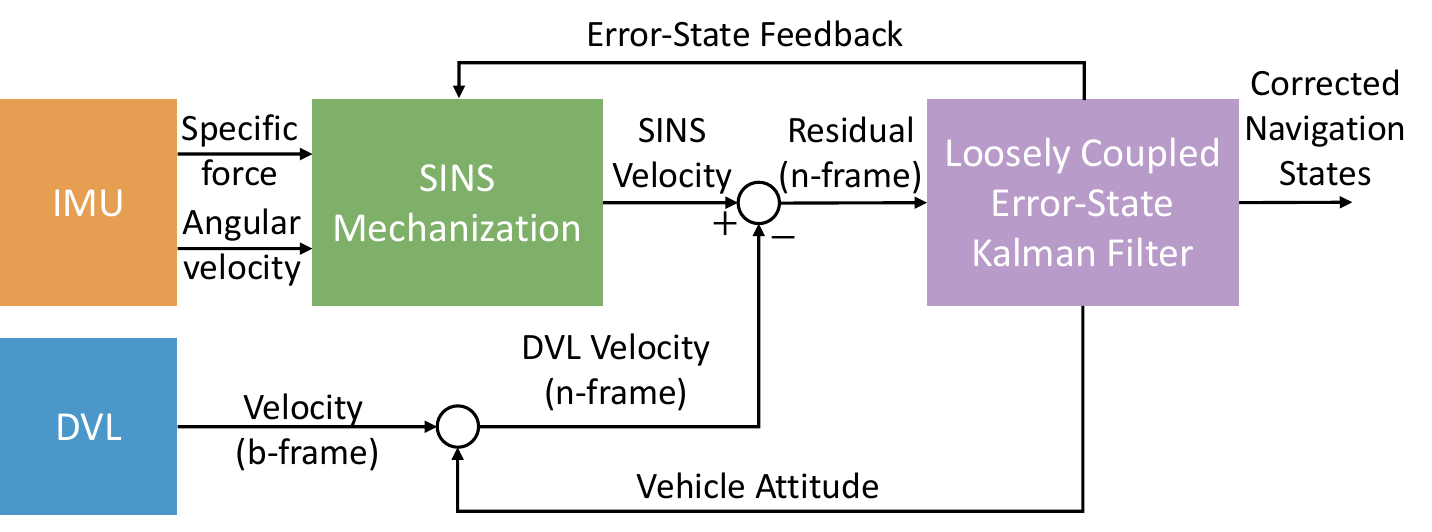}
  \caption{Loosely coupled SINS/DVL velocity-update signal flow considered for attitude-uncertain covariance propagation.}
  \label{figSystemFramework}
\end{figure}

This section summarizes the loosely coupled SINS/DVL navigation model used as the basis for the proposed covariance propagation method.
As shown in Fig.~\ref{figSystemFramework}, the considered system consists of an inertial measurement unit (IMU) and a DVL.
The IMU provides angular-rate and specific-force measurements for SINS mechanization, whereas the DVL provides body-frame velocity measurements for the loosely coupled velocity update.

The linearized error-state model and the corresponding measurement equation of the loosely coupled SINS/DVL integrated navigation system can be written as

\begin{equation}
  \begin{cases}
    \dot{\boldsymbol{X}}=\boldsymbol{F}\boldsymbol{X}+\boldsymbol{G}\boldsymbol{W}\\
    \boldsymbol{Z}=\boldsymbol{H}\boldsymbol{X}+\boldsymbol{N}
  \end{cases}
\end{equation}

where $\boldsymbol{X}$ is the error-state vector, $\boldsymbol{F}$ is the system dynamics matrix, $\boldsymbol{G}$ is the noise distribution matrix, $\boldsymbol{W}$ is the process-noise vector, $\boldsymbol{Z}$ is the observation vector, $\boldsymbol{H}$ is the observation matrix, and $\boldsymbol{N}$ is the measurement-noise vector \cite{grovesPrinciplesGNSSInertial2013, niuDevelopmentEvaluationGNSS2015}.

The 21-dimensional error-state vector $\boldsymbol{X}$ is defined as

\begin{equation}
\boldsymbol{X}=\left[ \begin{matrix}
	\left( \delta \boldsymbol{r}^n \right) ^{\top}&		\left( \delta \boldsymbol{v}^n \right) ^{\top}&		\boldsymbol{\phi }^{\top}&		\boldsymbol{b}_{g}^{\top}&		\boldsymbol{b}_{a}^{\top}&		\boldsymbol{s}_{g}^{\top}&		\boldsymbol{s}_{a}^{\top}\\
\end{matrix} \right] ^{\top}
\label{eq_stateVector}
\end{equation}
where $\delta \boldsymbol{r}^n$ and $\delta \boldsymbol{v}^n$ denote the position and velocity errors in the $n$-frame, respectively. The attitude-error vector is denoted by $\boldsymbol{\phi}$. The gyroscope and accelerometer bias errors are denoted by $\boldsymbol{b}_{g}$ and $\boldsymbol{b}_{a}$, respectively. The corresponding scale-factor errors are denoted by $\boldsymbol{s}_{g}$ and $\boldsymbol{s}_{a}$.

For the loosely coupled velocity update, the observation vector is constructed from the difference between the SINS-predicted velocity of the DVL phase center and the DVL-derived velocity after projection into the $n$-frame:
\begin{equation}
  \label{eq:observationVector}
  \boldsymbol{Z}=\hat{\boldsymbol{v}}_{\mathrm{DVL}}^{n}-\tilde{\boldsymbol{v}}_{\mathrm{DVL}}^{n}
\end{equation}
where $\hat{\boldsymbol{v}}_{\mathrm{DVL}}^{n}$ represents the SINS-predicted velocity of the DVL phase center, and $\tilde{\boldsymbol{v}}_{\mathrm{DVL}}^{n}$ denotes the DVL-measured velocity transformed from the $b$-frame to the $n$-frame.
This velocity-difference observation provides the basis for the attitude-error-aware DVL observation formulation and covariance propagation developed in Section~\ref{secProposedMethod}.

\section{DVL Observation and Covariance Propagation Under Attitude Uncertainty}
\label{secProposedMethod}

This section develops the DVL velocity observation model under attitude uncertainty and derives the corresponding measurement-noise covariance propagation.
The objective is to construct the navigation-frame DVL covariance used in the loosely coupled velocity update, while accounting for the uncertainty of the attitude used in the body-to-navigation-frame projection.
Hereafter, the IMU reference center is used as the SINS mechanization point unless otherwise specified.

\subsection{Attitude-Error-Aware DVL Measurement Equation}

The DVL-to-IMU installation angles and the lever arm are assumed to have been calibrated \cite{huangGNSSaidedInstallationError2025}.
After this calibration, the DVL velocity can be represented in the $b$-frame as
\begin{equation}
\boldsymbol{v}_{\mathrm{DVL}}^{b}=\left[ \begin{matrix}
	v_{x}^{b}&		v_{y}^{b}&		v_{z}^{b}\\
\end{matrix} \right] ^{\top}
\end{equation}
where $v_{x}^{b}$, $v_{y}^{b}$, and $v_{z}^{b}$ are the velocity components along the front-right-bottom (FRB) body-frame axes, respectively.

As shown in \eqref{eq_stateVector}, the velocity-error state is defined in the navigation frame.
Therefore, the DVL velocity measurement used in the loosely coupled update must be projected from the $b$-frame to the $n$-frame.
When the attitude used for this projection is uncertain, both the projected velocity and its associated measurement covariance are affected.

Let $\boldsymbol{C}_{b}^{n}$ denote the true direction cosine matrix (DCM) from the $b$-frame to the true $n$-frame, and let $\hat{\boldsymbol{C}}_{b}^{n^{\prime}}$ denote the SINS-computed DCM from the $b$-frame to the computed $n^{\prime}$-frame.
The DVL velocity projected by the true attitude is
\begin{equation}
  \bar{\boldsymbol{v}}_{\mathrm{DVL}}^{n}
  =
  \boldsymbol{C}_{b}^{n}\boldsymbol{v}_{\mathrm{DVL}}^{b}.
\end{equation}
Here, $n^{\prime}$ denotes the SINS-computed navigation frame, and $\boldsymbol{C}_{n}^{n^{\prime}}$ transforms vectors from the true $n$-frame to the computed $n^{\prime}$-frame.
Under the first-order small-angle approximation used in recent body-frame SINS/DVL attitude-error modeling \cite{gaoEulerAngleError2025},
\begin{equation}
  \hat{\boldsymbol{C}}_{b}^{n^{\prime}}
  =
  \boldsymbol{C}_{n}^{n^{\prime}}\boldsymbol{C}_{b}^{n}
  \approx
  \left[\boldsymbol{I}-\skew{\boldsymbol{\phi}}\right]\boldsymbol{C}_{b}^{n}
\end{equation}
where $\boldsymbol{I}$ is the identity matrix, $\boldsymbol{\phi}$ is the attitude-error vector, and $\skew{\boldsymbol{a}}$ denotes the skew-symmetric matrix satisfying $\skew{\boldsymbol{a}}\boldsymbol{b}=\boldsymbol{a}\times\boldsymbol{b}$.
For notational compactness, the prime on the computed navigation frame is omitted in the following filter equations.
Unless otherwise stated, all projected measurement quantities are expressed in the SINS-computed navigation frame.

The lever arm $\boldsymbol{l}^b$ is defined as the vector from the IMU reference center to the DVL phase center, expressed in the $b$-frame.
With lever-arm kinematics included and other measurement errors omitted, the IMU-center velocity and DVL phase-center velocity satisfy
\begin{equation}
  \boldsymbol{v}_{\mathrm{IMU}}^{n}
  =
  \bar{\boldsymbol{v}}_{\mathrm{DVL}}^{n}
  +\skew{\boldsymbol{\omega}_{in}^{n}}\boldsymbol{C}_{b}^{n}\boldsymbol{l}^b
  +\boldsymbol{C}_{b}^{n}\skew{\boldsymbol{l}^b}\boldsymbol{\omega}_{ib}^{b},
  \label{eq_velIMU}
\end{equation}
where $\boldsymbol{\omega}_{in}^{n}$ is the angular rate of the $n$-frame with respect to the inertial frame, expressed in the $n$-frame, and $\boldsymbol{\omega}_{ib}^{b}$ is the gyroscope output in the $b$-frame.

To derive the linearized velocity observation, the following error perturbations are defined:
\begin{equation}
  \begin{cases}
    \hat{\boldsymbol{v}}_{\mathrm{IMU}}^{n}=\boldsymbol{v}_{\mathrm{IMU}}^{n}+\delta \boldsymbol{v}^n\\
    \tilde{\boldsymbol{v}}_{\mathrm{DVL}}^{b}=\boldsymbol{v}_{\mathrm{DVL}}^{b}+\boldsymbol{n}_{v}^{b}\\
    \hat{\boldsymbol{\omega}}_{ib}^{b}=\boldsymbol{\omega }_{ib}^{b}+\delta \boldsymbol{\omega }_{ib}^{b}\\
    \delta \boldsymbol{\omega }_{ib}^{b}=\boldsymbol{b}_g+\mathrm{diag}\left( \boldsymbol{\omega }_{ib}^{b} \right) \boldsymbol{s}_g+\boldsymbol{\omega }_{\phi}
  \end{cases}
  \label{eq_error}
\end{equation}
where $\delta \boldsymbol{v}^n$, $\boldsymbol{n}_{v}^{b}$, $\boldsymbol{b}_g$, $\boldsymbol{s}_g$, and $\boldsymbol{\omega}_{\phi}$ denote the velocity error, DVL measurement noise, gyroscope bias error, gyroscope scale-factor error, and gyroscope white noise, respectively.

The SINS-predicted and DVL-measured velocities of the DVL phase center are defined as
\begin{equation}
  \label{eq:v_hat_DVL_n}
  \hat{\boldsymbol{v}}_{\mathrm{DVL}}^{n}
  =
  \hat{\boldsymbol{v}}_{\mathrm{IMU}}^{n}
  -
  \skew{\boldsymbol{\omega}_{in}^{n}}\hat{\boldsymbol{C}}_{b}^{n}\boldsymbol{l}^b
  -
  \hat{\boldsymbol{C}}_{b}^{n}\skew{\boldsymbol{l}^b}\hat{\boldsymbol{\omega}}_{ib}^{b},
\end{equation}
\begin{equation}
  \label{eq:v_tilde_DVL_n}
  \tilde{\boldsymbol{v}}_{\mathrm{DVL}}^{n}
  =
  \hat{\boldsymbol{C}}_{b}^{n}\tilde{\boldsymbol{v}}_{\mathrm{DVL}}^{b}.
\end{equation}
After substituting \eqref{eq_error} into \eqref{eq:v_hat_DVL_n} and \eqref{eq:v_tilde_DVL_n}, applying first-order linearization, and retaining the terms that contribute to the first-order measurement mean, the velocity residual becomes
\begin{equation}
  \label{eq:v_hat_DVL_n_expanded}
  \begin{aligned}
    \boldsymbol{Z}_v
    & = \hat{\boldsymbol{v}}_{\mathrm{DVL}}^{n}-\tilde{\boldsymbol{v}}_{\mathrm{DVL}}^{n} \\
    & \approx
    \delta \boldsymbol{v}^n
    +\skew{\boldsymbol{\omega}_{in}^{n}}\skew{\boldsymbol{\phi}}\boldsymbol{C}_{b}^{n}\boldsymbol{l}^b
    -\boldsymbol{C}_{b}^{n}\skew{\boldsymbol{l}^b}\delta \boldsymbol{\omega}_{ib}^b \\
    & \quad
    +\skew{\boldsymbol{\phi}}\boldsymbol{C}_{b}^{n}\skew{\boldsymbol{l}^b}\boldsymbol{\omega}_{ib}^{b}
    -\boldsymbol{C}_{b}^{n}\boldsymbol{n}_{v}^{b}
    +\skew{\boldsymbol{\phi}}\boldsymbol{C}_{b}^{n}\boldsymbol{v}_{\mathrm{DVL}}^{b}.
  \end{aligned}
\end{equation}
The bilinear term $\skew{\boldsymbol{\phi}}\boldsymbol{C}_{b}^{n}\boldsymbol{n}_{v}^{b}$ is omitted from the first-order measurement mean.
However, its second-order statistical contribution is retained in the covariance propagation.
The corresponding effective DVL noise is
\begin{equation}
  \boldsymbol{n}_{v,\mathrm{eff}}^{n}
  =
  -\boldsymbol{C}_{b}^{n}\boldsymbol{n}_{v}^{b}
  +
  \skew{\boldsymbol{\phi}}\boldsymbol{C}_{b}^{n}\boldsymbol{n}_{v}^{b}.
  \label{eq_noise_eff}
\end{equation}

The deterministic part of the velocity residual can then be written in the following linear form:
\begin{equation}
  \boldsymbol{Z}_v=\boldsymbol{H}_v\boldsymbol{X}+\boldsymbol{N}_v,
\end{equation}
where
\begin{equation}
  \boldsymbol{H}_v
  =
  \left[
  \boldsymbol{0}_{3}\;
  \boldsymbol{I}_3\;
  \boldsymbol{H}_{\phi}\;
  \boldsymbol{H}_{bg}\;
  \boldsymbol{0}_{3}\;
  \boldsymbol{H}_{sg}\;
  \boldsymbol{0}_{3}
  \right],
\end{equation}
\begin{equation}
  \begin{aligned}
    \boldsymbol{H}_{\phi}
    &=
    -\skew{\boldsymbol{\omega }_{in}^{n}}\skew{\boldsymbol{C}_{b}^{n}\boldsymbol{l}^b}
    -\skew{\boldsymbol{C}_{b}^{n}\skew{\boldsymbol{l}^b}\boldsymbol{\omega }_{ib}^{b}}\\
    &\quad
    -\skew{\boldsymbol{C}_{b}^{n}\boldsymbol{v}_{\mathrm{DVL}}^{b}},\\
    \boldsymbol{H}_{bg}
    &=
    -\boldsymbol{C}_{b}^{n}\skew{\boldsymbol{l}^b},\\
    \boldsymbol{H}_{sg}
    &=
    -\boldsymbol{C}_{b}^{n}\skew{\boldsymbol{l}^b}
    \mathrm{diag}\left( \boldsymbol{\omega }_{ib}^{b} \right).
  \end{aligned}
\end{equation}
In implementation, the true velocity $\boldsymbol{v}_{\mathrm{DVL}}^{b}$ in $\boldsymbol{H}_{\phi}$ is unavailable and is replaced by the current DVL measurement $\tilde{\boldsymbol{v}}_{\mathrm{DVL}}^{b}$.
This substitution introduces a stochastic product of DVL velocity noise and attitude error.
The product is excluded from the deterministic observation matrix and treated instead as part of the effective measurement noise in \eqref{eq_noise_eff}, thereby retaining its statistical contribution in the covariance propagation.

The first-order measurement-noise term in the linear residual is
\begin{equation}
  \boldsymbol{N}_v
  =
  -\boldsymbol{C}_{b}^{n}\boldsymbol{n}_{v}^{b}
  -\boldsymbol{C}_{b}^{n}\skew{\boldsymbol{l}^{b}}\boldsymbol{\omega}_{\phi}.
  \label{eq_noise_v}
\end{equation}
Here, $\boldsymbol{0}_{3}$ is a $3\times 3$ zero matrix and $\boldsymbol{I}_3$ is a $3\times 3$ identity matrix.

\subsection{DVL Measurement-Noise Covariance Propagation}

DVL bottom-track velocity is inferred from acoustic Doppler measurements, and its error characteristics may depend on beam geometry, seabed scattering, vehicle attitude, and bottom-lock conditions.
These factors may lead to axis-dependent and cross-correlated velocity errors.
Therefore, the DVL velocity noise is represented in the $b$-frame by a covariance matrix rather than by directly rotating a vector of standard deviations \cite{millerAutonomousUnderwaterVehicle2010}:
\begin{equation}
  \boldsymbol{\varSigma}_{v}^{b}
  =
  \begin{bmatrix}
    \sigma_{x}^{2} & \sigma_{xy} & \sigma_{xz}\\
    \sigma_{yx} & \sigma_{y}^{2} & \sigma_{yz}\\
    \sigma_{zx} & \sigma_{zy} & \sigma_{z}^{2}
  \end{bmatrix},
\end{equation}
where the off-diagonal terms allow possible cross-correlations among the DVL velocity errors.
In practice, the adopted body-frame DVL covariance can be obtained from manufacturer velocity-accuracy specifications, laboratory or field calibration, or DVL-reported quality information, such as standard-deviation estimates and validity indicators \cite{teledynePathfinderDVLGuide2025}.
When only axis-wise standard deviations are available, as in the specification-based setting used in the reported experiments, $\boldsymbol{\varSigma}_{v}^{b}$ is approximated by $\mathrm{diag}(\sigma_{x}^{2},\sigma_{y}^{2},\sigma_{z}^{2})$.
ACP does not estimate the intrinsic DVL noise model. It propagates the adopted body-frame covariance through the attitude-uncertain projection.
Estimating a time-varying seabed- or beam-quality-dependent DVL noise covariance is outside the scope of this work.
Compared with innovation- or residual-driven covariance tuning, ACP does not introduce an additional residual-driven adaptation gain to scale the attitude-induced covariance term.
The magnitude of this term is determined by the adopted body-frame DVL covariance and the predicted attitude-error covariance already available from the filter prediction.

Before attitude uncertainty is considered, the nominal matrix-level transformation propagates the body-frame DVL covariance into the navigation frame as
\begin{equation}
  \boldsymbol{\varSigma}_{v,0}^{n}
  =
  \boldsymbol{C}_{b}^{n}\boldsymbol{\varSigma}_{v}^{b}
  \left(\boldsymbol{C}_{b}^{n}\right)^{\top}.
  \label{eq_covNorm}
\end{equation}
This standard covariance transformation is required because the DVL noise is specified in the body frame, whereas the loosely coupled velocity residual is formed in the navigation frame.
However, \eqref{eq_covNorm} does not account for uncertainty in the attitude used for the body-to-navigation-frame projection.
With the prime on the computed navigation frame omitted, $\boldsymbol{C}_{b}^{n}$ in the implemented covariance construction denotes the current nominal DCM computed by the SINS.
From \eqref{eq_noise_eff}, the attitude-aware effective DVL noise is
\begin{equation}
  \boldsymbol{n}_{v,\mathrm{eff}}^{n}
  =
  -\boldsymbol{C}_{b}^{n}\boldsymbol{n}_{v}^{b}
  +
  \skew{\boldsymbol{\phi}}\boldsymbol{C}_{b}^{n}\boldsymbol{n}_{v}^{b}.
\end{equation}

Let $\boldsymbol{A}\triangleq\boldsymbol{\varSigma}_{v,0}^{n}$.
The attitude error $\boldsymbol{\phi}$ and DVL noise $\boldsymbol{n}_{v}^{b}$ are assumed to be zero-mean and mutually independent, conditioned on the adopted body-frame DVL noise model.
Under this assumption and the first-order small-angle perturbation model, the cross terms vanish.
This assumption is used to construct the covariance in the filter error state. It does not preclude attitude- or environment-dependent changes in the prescribed DVL noise model.
Therefore, the DVL measurement-noise covariance in the $n$-frame becomes
\begin{equation}
  \begin{aligned}
    \boldsymbol{\varSigma}_{v}^{n}
    &=
    \mathbb{E}\!\left[
    \boldsymbol{n}_{v,\mathrm{eff}}^{n}
    \left(\boldsymbol{n}_{v,\mathrm{eff}}^{n}\right)^{\top}
    \right] \\
    &=
    \boldsymbol{A}
    +
    \mathbb{E}\!\left[
      \skew{\boldsymbol{\phi}}\boldsymbol{A}\skew{\boldsymbol{\phi}}^{\top}
    \right].
  \end{aligned}
  \label{eq_cov_with_phi_expect}
\end{equation}

The attitude-error covariance $\boldsymbol{P}_{\phi\phi}^{-}$ is extracted from the predicted state covariance before the DVL measurement update.
The above approximation is intended for the post-alignment navigation stage, where the predicted attitude-error covariance remains consistent with the small-angle linearization.
Its applicable range depends on the system and required accuracy rather than on a universal angular threshold.
For severe initial misalignment, large attitude uncertainty, or aggressive maneuvers, higher-order attitude-error terms may become non-negligible. In such cases, coarse alignment, nonlinear attitude-uncertainty propagation, or higher-order covariance propagation should be used before applying the present ACP method.

By expressing $\skew{\boldsymbol{\phi}}=\sum_{i=1}^{3}\phi_i\boldsymbol{S}_i$, where $\boldsymbol{S}_i$ are the constant basis matrices of the skew-symmetric representation, the expectation term has the following closed form:
\begin{equation}
  \mathbb{E}\!\left[
    \skew{\boldsymbol{\phi}}\boldsymbol{A}\skew{\boldsymbol{\phi}}^{\top}
  \right]
  =
  \sum_{i=1}^{3}\sum_{j=1}^{3}
  \left(\boldsymbol{P}_{\phi\phi}^{-}\right)_{ij}
  \boldsymbol{S}_{i}\boldsymbol{A}\boldsymbol{S}_{j}^{\top}.
  \label{eq_phi_closed_form}
\end{equation}

For interpretation, the magnitude of the closed-form term in \eqref{eq_phi_closed_form} can be compared with the nominal transformed DVL covariance.
Accordingly, the attitude-induced covariance ratio (AICR) is defined as the following diagnostic quantity:
\begin{equation}
  \eta_{\phi,k}
  =
  \frac{
  \mathrm{tr}\!\left(\Delta\boldsymbol{\varSigma}_{v,\phi,k}^{n}\right)
  }{
  \mathrm{tr}\!\left(\boldsymbol{\varSigma}_{v,0,k}^{n}\right)
  }
  \times 100\%,
  \label{eq:aicr}
\end{equation}
where $\Delta\boldsymbol{\varSigma}_{v,\phi,k}^{n}$ denotes the attitude-induced covariance term computed by \eqref{eq_phi_closed_form}.
This ratio is not used by the filter and introduces no additional tuning parameter. It only normalizes the attitude-induced covariance contribution for scale interpretation.
If $\boldsymbol{P}_{\phi\phi}=\sigma_{\phi}^{2}\boldsymbol{I}_{3}$, \eqref{eq:aicr} reduces to
\begin{equation}
  \eta_{\phi}\left(\sigma_{\phi}\right)
  =
  2\sigma_{\phi}^{2}\times 100\%,
  \label{eq:aicr_isotropic}
\end{equation}
where $\sigma_{\phi}$ is expressed in radians.
For isotropic diagnostic inputs, attitude-error standard deviations of $1^{\circ}$, $2^{\circ}$, and $5^{\circ}$ correspond to AICR values of approximately $0.0609\%$, $0.2437\%$, and $1.5231\%$, respectively.

The attitude-error term in $\boldsymbol{H}_{\phi}$ models the deterministic perturbation of the projected DVL velocity, whereas the covariance propagation term models the second-order statistical effect of attitude uncertainty on the transformed DVL measurement noise.
Thus, the covariance propagation modifies the measurement-noise weighting and does not duplicate the attitude-error state update.

The DVL-related measurement-noise covariance used in the filter is therefore
\begin{equation}
  \boldsymbol{R}_{v,\mathrm{DVL}}
  =
  \boldsymbol{\varSigma}_{v}^{n}.
  \label{eq:Rv_DVL}
\end{equation}
Assuming that the DVL velocity noise and the gyroscope white noise are mutually independent, including the gyroscope white-noise term in \eqref{eq_noise_v} gives
\begin{equation}
  \label{eq:Rv_final}
  \boldsymbol{R}_{v}
  =
  \boldsymbol{R}_{v,\mathrm{DVL}}
  +
  \boldsymbol{C}_{b}^{n}
  \skew{\boldsymbol{l}^{b}}
  \boldsymbol{Q}_{\omega}
  \skew{\boldsymbol{l}^{b}}^{\top}
  \left(\boldsymbol{C}_{b}^{n}\right)^{\top},
\end{equation}
where $\boldsymbol{Q}_{\omega}=\mathbb{E}\left[\boldsymbol{\omega}_{\phi}\boldsymbol{\omega}_{\phi}^{\top}\right]$ is the discrete covariance matrix of the gyroscope white noise at the DVL update epoch.

\subsection{Implementation Procedure}
\label{secImplementationProcedure}

The proposed velocity update is implemented at each DVL measurement epoch after the SINS time propagation and covariance prediction.
The main operations are summarized in Algorithm~\ref{algACPFlow}.
The procedure first constructs the attitude-error-aware velocity residual and observation matrix, and then builds the measurement-noise covariance by propagating the body-frame DVL covariance through the attitude-uncertain projection.

\begin{algorithm}[H]
  \caption{Attitude-Error-Aware DVL Update With ACP}
  \label{algACPFlow}
  \footnotesize
  \begin{algorithmic}[1]
    \REQUIRE Predicted SINS state and covariance $\boldsymbol{P}^{-}$, DVL measurement $\tilde{\boldsymbol{v}}_{\mathrm{DVL}}^{b}$, body-frame DVL covariance $\boldsymbol{\varSigma}_{v}^{b}$, lever arm $\boldsymbol{l}^{b}$, and gyroscope white-noise covariance $\boldsymbol{Q}_{\omega}$
    \ENSURE Updated navigation state and covariance, with constructed $\boldsymbol{Z}_{v}$, $\boldsymbol{H}_{v}$, and $\boldsymbol{R}_{v}$
    \STATE Project the DVL velocity to the navigation frame using the current SINS-computed attitude.
    \STATE Compute the SINS-predicted DVL phase-center velocity with the lever-arm correction.
    \STATE Form the velocity residual $\boldsymbol{Z}_{v}$ according to \eqref{eq:observationVector}.
    \STATE Assemble the observation matrix $\boldsymbol{H}_{v}$ using \eqref{eq:v_hat_DVL_n_expanded}--\eqref{eq_noise_v}, including the attitude-, gyroscope-bias-, and gyroscope-scale-factor-related blocks.
    \STATE Extract the predicted attitude-error covariance $\boldsymbol{P}_{\phi\phi}^{-}$ from $\boldsymbol{P}^{-}$.
    \STATE Compute the nominal transformed covariance $\boldsymbol{A}$ using the current SINS-computed nominal attitude, as in \eqref{eq_covNorm}.
    \STATE Evaluate the attitude-induced covariance term using the closed form in \eqref{eq_phi_closed_form}.
    \STATE Construct $\boldsymbol{R}_{v,\mathrm{DVL}}$ from the attitude-aware covariance in \eqref{eq_cov_with_phi_expect} and \eqref{eq:Rv_DVL}.
    \STATE Add the gyroscope white-noise contribution to obtain $\boldsymbol{R}_{v}$ according to \eqref{eq:Rv_final}.
    \STATE Perform the Kalman measurement update with $\boldsymbol{Z}_{v}$, $\boldsymbol{H}_{v}$, and $\boldsymbol{R}_{v}$.
    \STATE Feed back the estimated error state to correct the SINS navigation state and sensor-error states.
  \end{algorithmic}
\end{algorithm}
\FloatBarrier

\section{Results and Discussion}
\label{secResultDiscussion}

This section evaluates the complete ACP method, which combines the attitude-error-aware observation model with analytical covariance propagation, through simulation and a controlled surface-water field experiment.
The simulation compares conventional SINS/DVL, SINS/DVL with Sage-Husa adaptive covariance estimation, and SINS/DVL with ACP. The field experiment additionally includes SINS/DVL with a variational Bayesian adaptive Kalman filter (VBAKF) as a second adaptive-covariance benchmark \cite{wangNovelSINSDVL2024}.
Because the observation model and covariance propagation are coupled within ACP, the reported improvements are attributed to the complete method rather than to either component individually.

For the SINS/DVL (Sage-Husa) benchmark, the same nominal DVL observation model and DVL noise inputs as the conventional SINS/DVL baseline are used.
The adaptive covariance is initialized from the nominal transformed DVL measurement covariance and subsequently updated as a diagonal matrix using innovation statistics.
Specifically, with update gain \(d=0.02\), the instantaneous diagonal estimate is obtained from the squared innovation after subtraction of the predicted state-projection covariance. It is then clipped to the configured lower bounds, smoothed with the previous estimate, and limited by the configured standard-deviation bounds.
These constraints keep the covariance diagonal and positive during the update.

\subsection{Simulation}
\label{secSimulation}

\begin{figure}[!t]
  \centering
  \includegraphics[width=\linewidth]{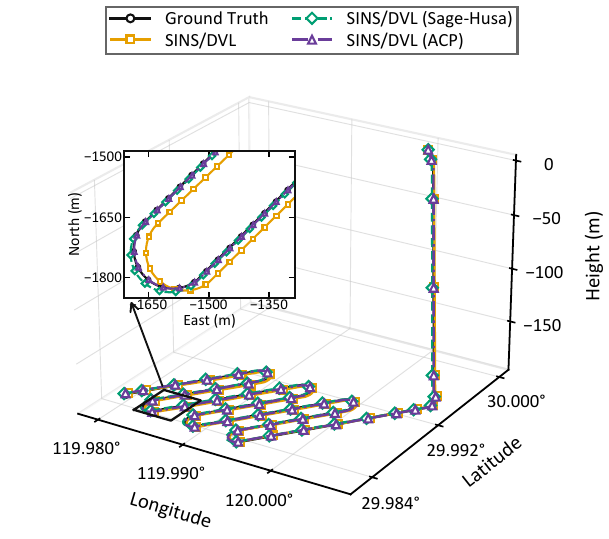}
  \caption{Simulated trajectory obtained with the compared SINS/DVL methods.}
  \label{figSimTrace}
\end{figure}

The simulation evaluates the compared covariance-handling strategies over a trajectory with prescribed changes in depth, speed, and heading.
As shown in Fig.~\ref{figSimTrace}, the trajectory lasts 4000 s, including 600 s of initial alignment and 3400 s of diving, acceleration, deceleration, and coordinated turns.
The initial latitude and longitude are 30.0 deg and 120.0 deg, respectively, and the initial depth is 0 m.
The sensor and sampling specifications used in the simulation are summarized in Table~\ref{tab_simSpec}.

\begin{table}
  \centering
    \caption{Simulation specifications.}
    \label{tab_simSpec}
    \begin{tabular}{l l l}
      \toprule
      Equipment & Specification & Accuracy \\
      \midrule
      Gyroscope & Constant bias & 0.01 deg/h \\
        & Random walk & 0.01 deg/$\sqrt{\mathrm{h}}$ \\
        & Scale factor error & 100 ppm \\
        &  Measurement frequency & 100 Hz \\
      Accelerometer & Constant bias & 50  $\mu$g \\
        & Random walk & 10 $\mu$g/$\sqrt{\mathrm{Hz}}$ \\
        & Scale factor error & 100 ppm \\
        &  Measurement frequency & 100 Hz \\
      DVL & Velocity error & 1.15\% of speed \\
        &  Measurement frequency & 2 Hz \\
      \bottomrule
    \end{tabular}
\end{table}

Fig.~\ref{figSimTrace} compares the trajectories estimated by the three methods with the ground truth.
The Sage-Husa method serves as a residual-driven adaptive-covariance benchmark, whereas ACP constructs the transformed DVL covariance through the closed-form propagation in~\eqref{eq_phi_closed_form}.
The simulation therefore compares an empirical covariance-adjustment strategy with the proposed model-driven covariance construction under the same trajectory and sensor settings.

\begin{figure}[!t]
  \centering
  \includegraphics[width=\linewidth]{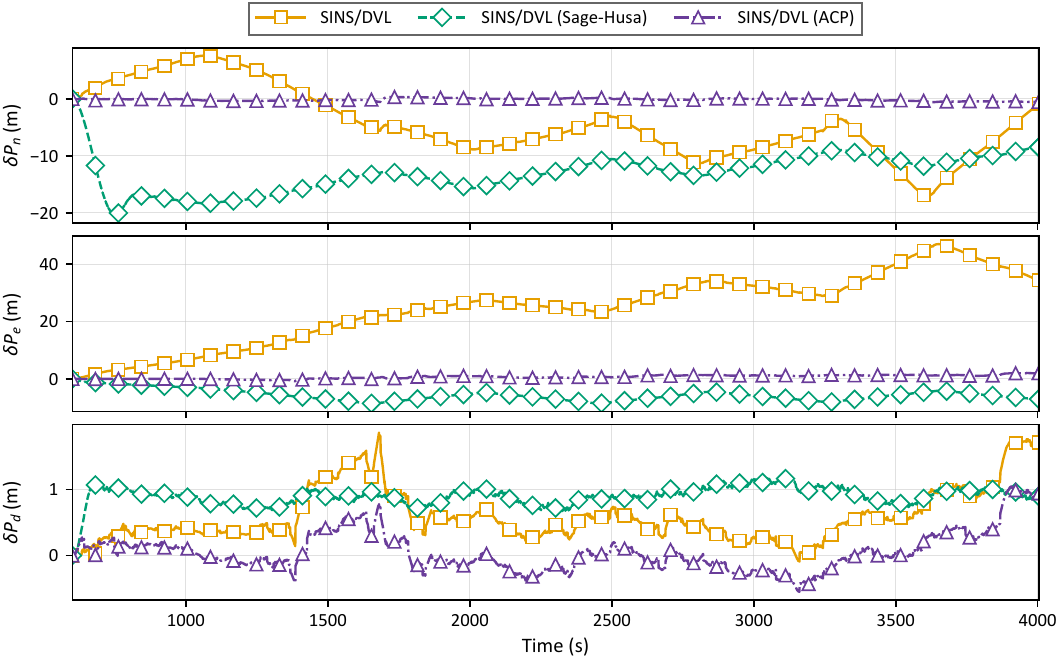}
  \caption{Simulated position errors of SINS/DVL, SINS/DVL (Sage-Husa), and SINS/DVL (ACP).}
  \label{figSimPosErr}
\end{figure}

\begin{figure}[!t]
  \centering
  \includegraphics[width=\linewidth]{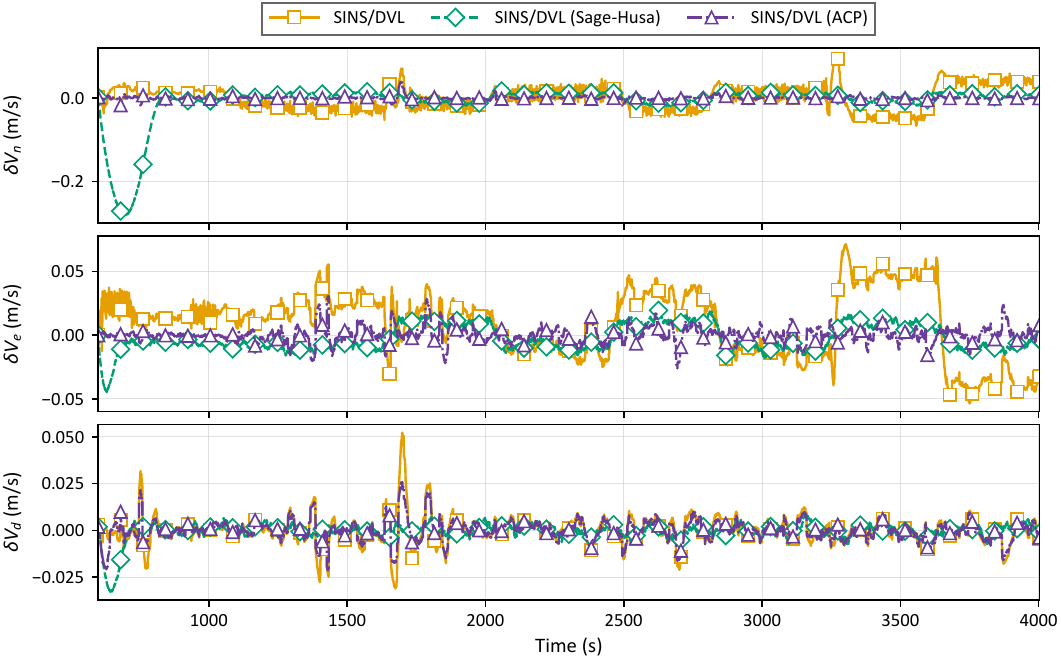}
  \caption{Simulated velocity errors of SINS/DVL, SINS/DVL (Sage-Husa), and SINS/DVL (ACP).}
  \label{figSimVelErr}
\end{figure}

\begin{figure}[!t]
  \centering
  \includegraphics[width=\linewidth]{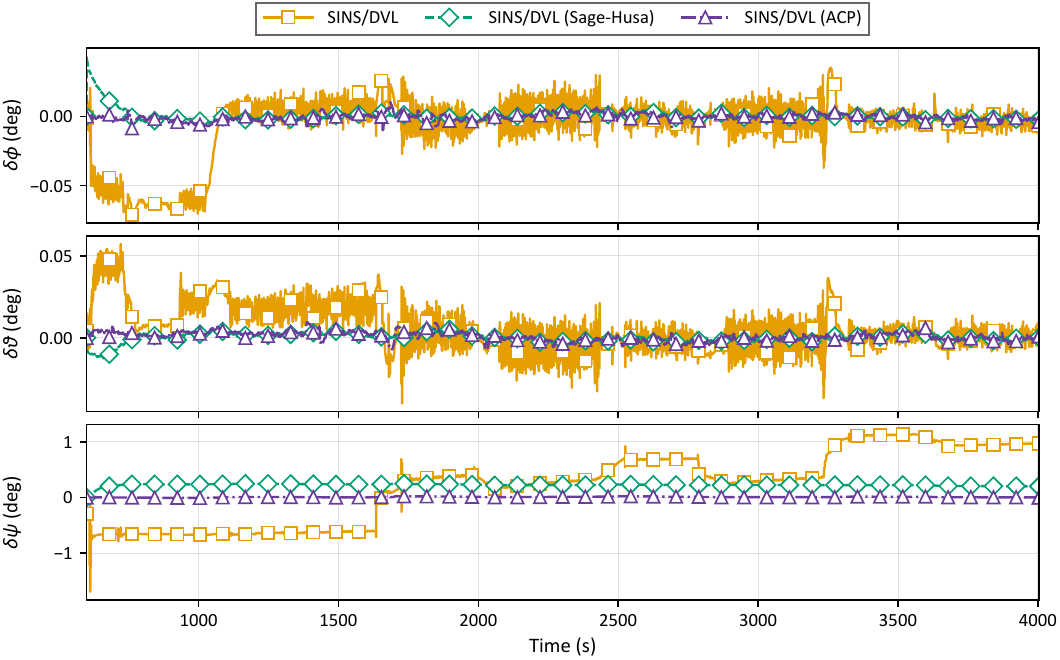}
  \caption{Simulated attitude errors of SINS/DVL, SINS/DVL (Sage-Husa), and SINS/DVL (ACP).}
  \label{figSimAttErr}
\end{figure}

The position, velocity, and attitude error histories are shown in Figs.~\ref{figSimPosErr}--\ref{figSimAttErr}, and the corresponding root-mean-square errors (RMSEs) and maximum errors are listed in Table~\ref{tab_simPosErr}.
The conventional SINS/DVL solution yields the largest position and attitude errors among the three methods.
In this configuration, the DVL velocity is projected using the estimated attitude, whereas the conventional measurement covariance does not explicitly represent the uncertainty introduced by that projection.
The SINS/DVL (Sage-Husa) solution reduces the position RMSE from 28.38 m to 14.75 m and the attitude RMSE from 0.67 deg to 0.23 deg, indicating that innovation- or residual-driven covariance tuning reduces error growth in this configuration.
However, its velocity RMSE increases from 0.040 m/s to 0.051 m/s, indicating that innovation-based covariance adjustment does not necessarily improve all error channels simultaneously.

SINS/DVL (ACP) yields the lowest RMSE and maximum error for the reported position, velocity, and attitude metrics.
Using SINS/DVL (Sage-Husa) as the main baseline, SINS/DVL (ACP) reduces the simulated position RMSE from 14.75 m to 1.01 m and the maximum position error from 20.46 m to 2.38 m, corresponding to reductions of 93.2\% and 88.4\%, respectively.
The velocity RMSE decreases from 0.051 m/s to 0.009 m/s, and the attitude RMSE decreases from 0.23 deg to 0.01 deg.
For this simulation and parameterization, these results are obtained with the complete ACP method, which accounts for attitude error in both the projected velocity observation and its covariance-based weighting.

\begin{table}
  \centering
    \caption{RMSE and maximum errors of the compared methods in simulation.}
    \label{tab_simPosErr}
    \begin{tabular}{llccc}
      \toprule
      \multicolumn{2}{c}{Error Type} & \makecell{SINS/DVL} & \makecell{SINS/DVL\\(Sage-Husa)} & \makecell{SINS/DVL\\(ACP)} \\
      \midrule
      
      \multirow{2}{*}{Pos (m)} & RMSE & 28.38 & 14.75 & 1.01 \\
        & Max & 49.49 & 20.46 & 2.38 \\
      
      \multirow{2}{*}{Vel (m/s)} & RMSE & 0.040 & 0.051 & 0.009 \\
        & Max & 0.108 & 0.280 & 0.049 \\
      
      \multirow{2}{*}{Att (deg)} & RMSE & 0.67 & 0.23 & 0.01 \\
        & Max & 1.70 & 0.25 & 0.03 \\
      \bottomrule
    \end{tabular}
\end{table}

\begin{figure}
  \centering
  \includegraphics[width=\linewidth]{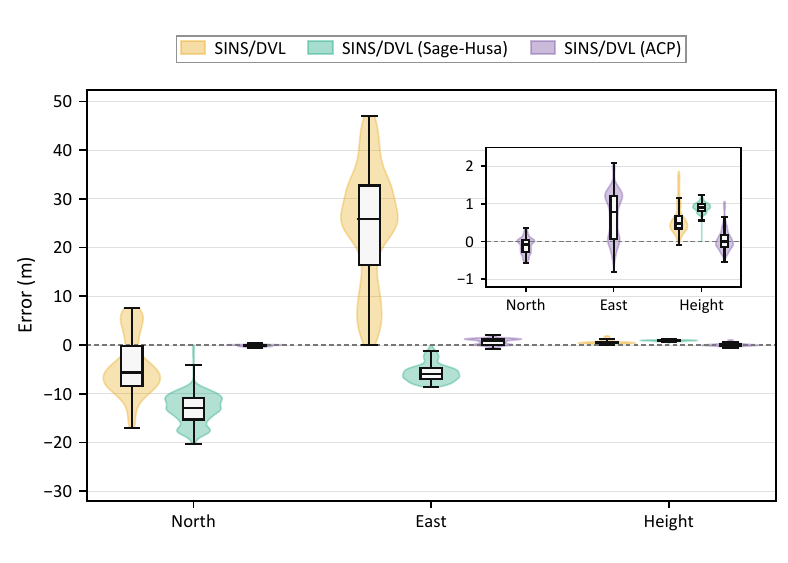}
  \caption{Distribution of simulated position errors for SINS/DVL, SINS/DVL (Sage-Husa), and SINS/DVL (ACP).}
  \label{figSimPosErrBox}
\end{figure}

Fig.~\ref{figSimPosErrBox} and Table~\ref{tab_simBoxStats} further characterize the component-wise position-error distributions using the median, interquartile range (IQR), and 95th-percentile absolute error.
Relative to SINS/DVL (Sage-Husa), SINS/DVL (ACP) reduces the northward and eastward IQRs from 4.46 m and 2.30 m to 0.31 m and 1.14 m, corresponding to reductions of 93.0\% and 50.5\%, respectively.
The signed medians in the northward and downward directions also move closer to zero, from -12.89 m and 0.91 m to -0.08 m and -0.01 m.
The eastward median of ACP is 0.78 m. Although nonzero, it remains substantially smaller than the conventional SINS/DVL median of 25.87 m.
For the distribution tails, ACP lowers the northward, eastward, and downward 95th-percentile absolute-error values to 0.48 m, 1.45 m, and 0.60 m, respectively.
The downward IQR of ACP is larger than that of SINS/DVL (Sage-Husa), but ACP still gives a smaller downward median bias and a lower 95th-percentile absolute error.
Overall, the distributional improvement is most pronounced in the horizontal components, whereas the downward component primarily shows reduced bias and tail error.

\begin{table}
  \centering
  \caption{Distribution statistics of the simulated component-wise position errors in Fig.~\ref{figSimPosErrBox}.}
  \label{tab_simBoxStats}
  \begin{tabular}{llccc}
    \toprule
    \multicolumn{2}{c}{Statistic} & \makecell{SINS/DVL} & \makecell{SINS/DVL\\(Sage-Husa)} & \makecell{SINS/DVL\\(ACP)} \\
    \midrule
    \multirow{3}{*}{$\delta P_N$} & Median & -5.67 & -12.89 & -0.08 \\
    & IQR & 8.21 & 4.46 & 0.31 \\
    & $P_{95}(|e|)$ & 13.34 & 18.10 & 0.48 \\
    \multirow{3}{*}{$\delta P_E$} & Median & 25.87 & -5.90 & 0.78 \\
    & IQR & 16.26 & 2.30 & 1.14 \\
    & $P_{95}(|e|)$ & 43.70 & 8.17 & 1.45 \\
    \multirow{3}{*}{$\delta P_D$} & Median & 0.48 & 0.91 & -0.01 \\
    & IQR & 0.33 & 0.17 & 0.31 \\
    & $P_{95}(|e|)$ & 1.47 & 1.10 & 0.60 \\
    \bottomrule
  \end{tabular}
\end{table}

\subsection{Field Experiment}
\label{secFieldExperiment}

\begin{figure}
  \centering
  \includegraphics[width=\linewidth]{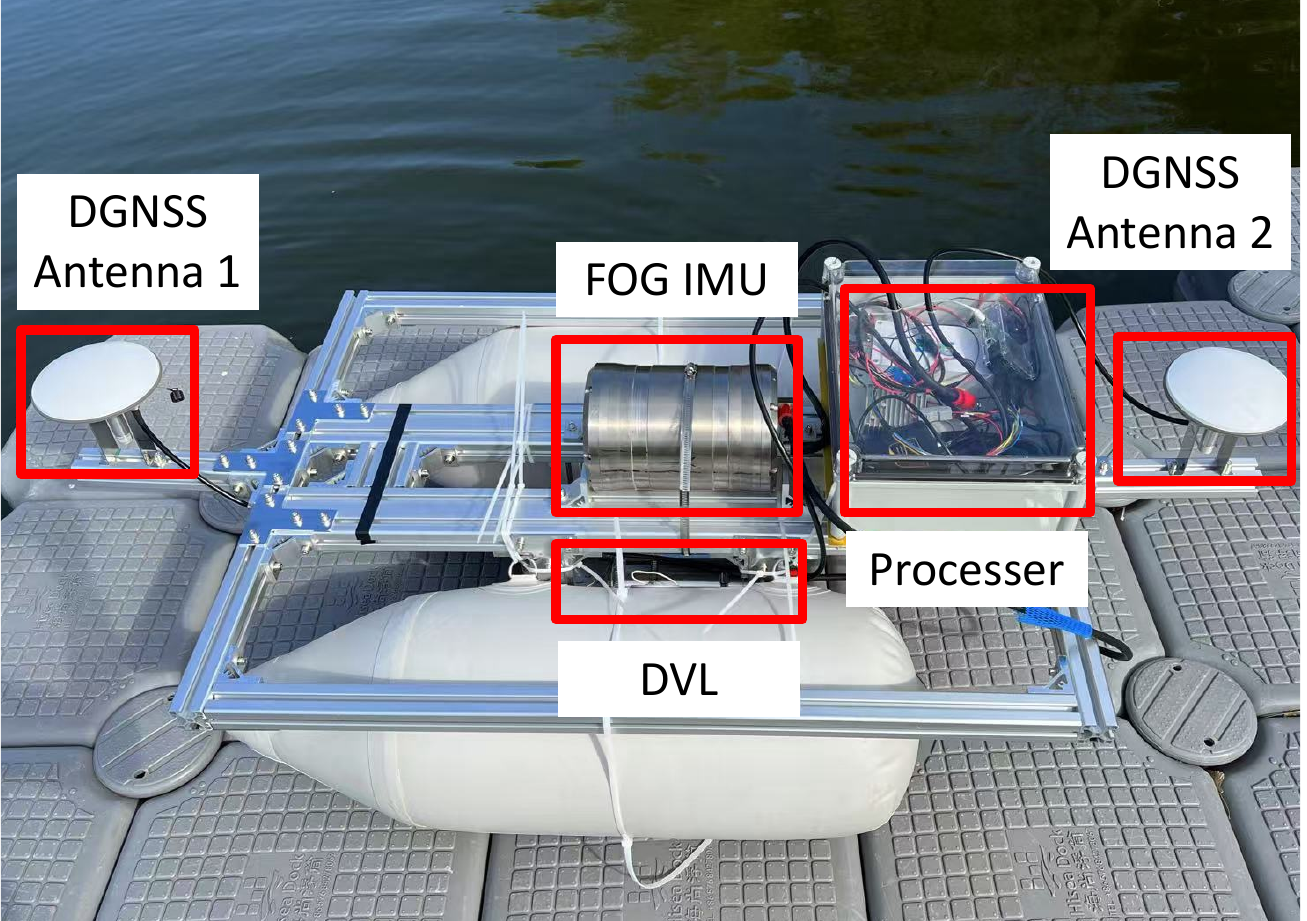}
  \caption{Surface-water experimental platform equipped with the SINS/DVL integrated navigation system and the DGNSS/CORS reference system.}
  \label{figExpPlatform}
\end{figure}

The field experiment comprised a controlled surface-water evaluation at Zhejiang University using real sensor data.
Because the platform operated near the water surface, differential GNSS (DGNSS) supported by a continuously operating reference station (CORS) provided centimeter-level reference positions.
The DGNSS/CORS trajectory serves only as an external performance reference and is not used as an absolute position update in the SINS/DVL filter.
The experimental platform is shown in Fig.~\ref{figExpPlatform}, and the sensor specifications are listed in Table~\ref{tab_expSpec}.
During the selected field-test interval, the DVL operated within its bottom-track range, and measurements were retained only when valid bottom-track velocity outputs were available.
The local water depth exceeded $2~\mathrm{m}$ and remained within the $0.2$--$89~\mathrm{m}$ bottom-track operating range specified for the selected DVL.
Accordingly, every DVL update used by the SINS/DVL filter was based on a valid bottom-track velocity output.

\begin{table}
  \centering
    \caption{Field experiment specifications.}
    \label{tab_expSpec}
    \begin{tabular}{l l l}
      \toprule
      Equipment & Specification & Accuracy \\
      \midrule
      Gyroscope & Constant bias & 0.03 deg/h \\
        & Random walk & 0.005 deg/$\sqrt{\mathrm{h}}$ \\
        & Scale factor error & 100 ppm \\
        &  Measurement frequency & 100 Hz \\
      Accelerometer & Constant bias & 300 $\mu$g \\
        & Random walk & 0.05 m/s/$\sqrt{\mathrm{h}}$ \\
        & Scale factor error & 200 ppm \\
        &  Measurement frequency & 100 Hz \\
      DVL & Velocity error & 1.15\% of speed \\
        &  Measurement frequency & 2 Hz \\
      DGNSS & Position (CORS) & 0.02 m \\
      \bottomrule
    \end{tabular}
\end{table}

\begin{figure}
  \centering
  \includegraphics[width=\linewidth]{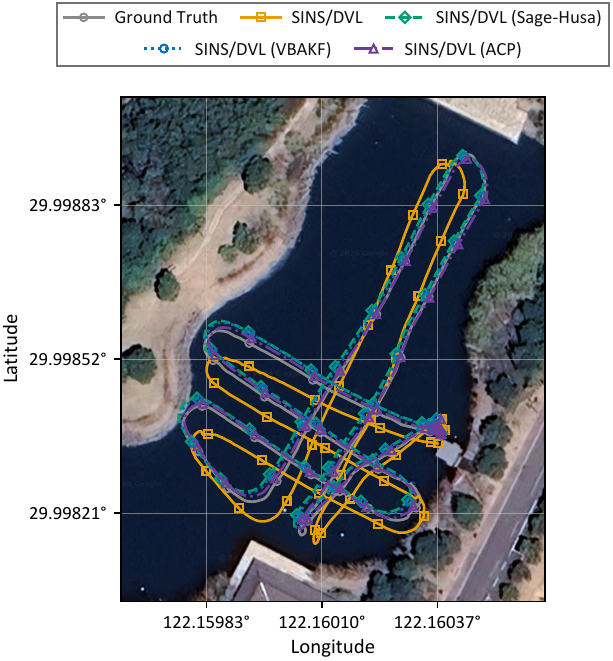}
  \caption{Surface-water experimental trajectory obtained with the compared SINS/DVL methods.}
  \label{figExpTrace}
\end{figure}

The experimental trajectory is shown in Fig.~\ref{figExpTrace}.
The analyzed record spans 1685 s, comprising 600 s of dynamic initial alignment and 1085 s of surface-water navigation.
All four methods are evaluated using the same field data record.
The trajectory plot provides a qualitative comparison of the field-test results, while the following position-error curves and statistical indicators quantify the component-wise differences among the compared methods.

\begin{figure}[!t]
  \centering
  \includegraphics[width=\linewidth]{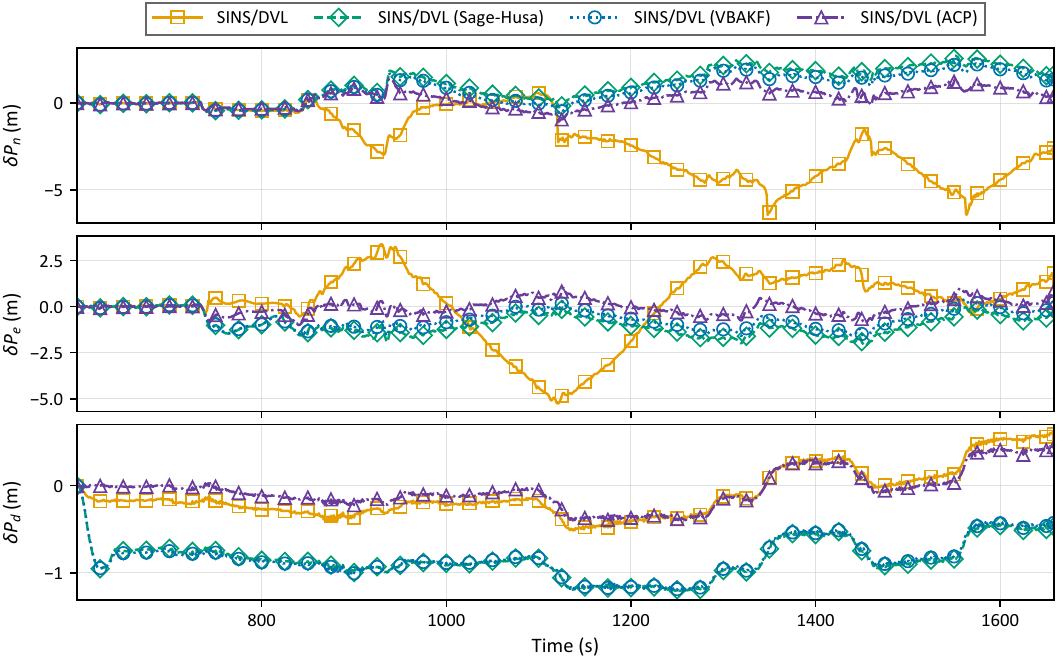}
  \caption{Field position errors of SINS/DVL, SINS/DVL (Sage-Husa), SINS/DVL (VBAKF), and SINS/DVL (ACP).}
  \label{figExpPosErr}
\end{figure}

Fig.~\ref{figExpPosErr} shows the north-east-down (NED) position errors obtained in the field experiment.
The conventional SINS/DVL solution accumulates the largest horizontal errors, with maximum northward and eastward component errors of 6.4527 m and 5.2547 m, respectively.
Both adaptive-covariance baselines reduce the horizontal errors relative to conventional SINS/DVL. However, the downward RMSE is 0.8686 m for SINS/DVL (Sage-Husa), while SINS/DVL (VBAKF) gives a similar value of 0.8650 m. Because VBAKF yields the lower RMSE of the two adaptive baselines in all three components, it is used as the stricter field benchmark.

As reported in Table~\ref{tab_expErrors}, relative to SINS/DVL (VBAKF), SINS/DVL (ACP) reduces the northward, eastward, and downward RMSE values from 1.1494 m, 0.8309 m, and 0.8650 m to 0.5923 m, 0.3735 m, and 0.2182 m, respectively. These changes correspond to reductions of 48.5\%, 55.0\%, and 74.8\%. The maximum component-wise errors decrease from 2.3800 m, 1.7185 m, and 1.2044 m to 1.4245 m, 1.0141 m, and 0.4549 m, corresponding to reductions of 40.1\%, 41.0\%, and 62.2\%, respectively. ACP therefore yields the lowest RMSE and maximum error in all three position components for this data set.

\begin{table}
  \centering
  \footnotesize
  \setlength{\tabcolsep}{2.5pt}
  \caption{RMSE and maximum position errors in the field experiment (m).}
  \label{tab_expErrors}
  \begin{tabular}{llcccc}
    \toprule
    \multicolumn{2}{c}{Error} & \makecell{SINS/DVL} & \makecell{SINS/DVL\\(Sage-Husa)} & \makecell{SINS/DVL\\(VBAKF)} & \makecell{SINS/DVL\\(ACP)} \\
    \midrule
    \multirow{2}{*}{$\delta P_N$} 
    & RMSE & 2.8266 & 1.3847 & 1.1494 & 0.5923 \\
    & Max & 6.4527 & 2.7336 & 2.3800 & 1.4245 \\
    \multirow{2}{*}{$\delta P_E$} 
    & RMSE & 1.8992 & 1.0918 & 0.8309 & 0.3735 \\
    & Max & 5.2547 & 2.1430 & 1.7185 & 1.0141 \\
    \multirow{2}{*}{$\delta P_D$} 
    & RMSE & 0.2951 & 0.8686 & 0.8650 & 0.2182 \\
    & Max & 0.6096 & 1.2200 & 1.2044 & 0.4549 \\
    \bottomrule
  \end{tabular}
\end{table}

\begin{figure}
  \centering
  \includegraphics[width=\linewidth]{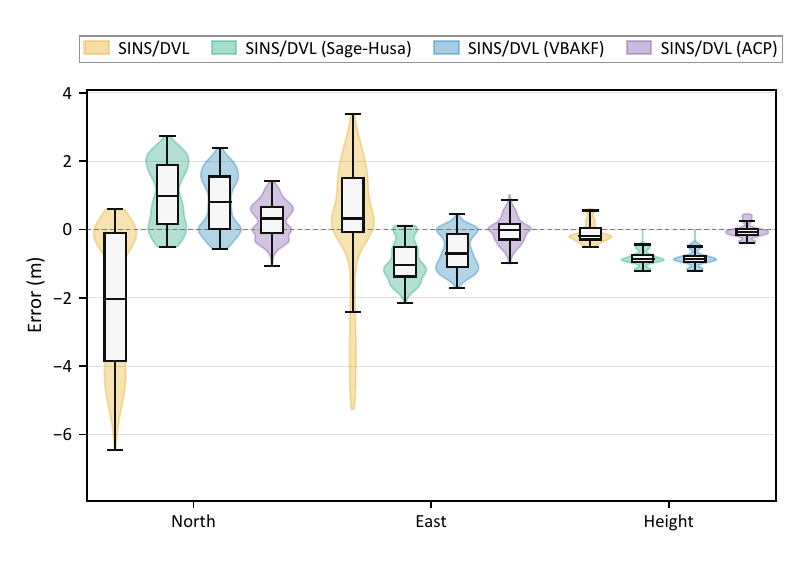}
  \caption{Distribution of field position errors for SINS/DVL, SINS/DVL (Sage-Husa), SINS/DVL (VBAKF), and SINS/DVL (ACP).}
  \label{figExpPosErrBox}
\end{figure}

Fig.~\ref{figExpPosErrBox} and Table~\ref{tab_fieldBoxStats} further characterize error dispersion and tail behavior beyond RMSE and maximum error. Relative to SINS/DVL (VBAKF), SINS/DVL (ACP) reduces the northward, eastward, and downward IQRs from 1.5492 m, 0.9576 m, and 0.1783 m to 0.7643 m, 0.4607 m, and 0.1656 m, corresponding to reductions of 50.7\%, 51.9\%, and 7.1\%, respectively. The corresponding 95th-percentile absolute errors decrease from 1.9955 m, 1.3958 m, and 1.1686 m to 1.1386 m, 0.7156 m, and 0.3980 m, reductions of 42.9\%, 48.7\%, and 65.9\%. The ACP medians are also closer to zero in all three components. For this data set and parameterization, these statistics indicate lower central dispersion and tail errors with ACP, although the downward IQR reduction relative to VBAKF is modest.

\begin{table}
  \centering
  \footnotesize
  \setlength{\tabcolsep}{2.5pt}
  \caption{Distribution statistics of the field component-wise position errors in Fig.~\ref{figExpPosErrBox}.}
  \label{tab_fieldBoxStats}
  \begin{tabular}{llcccc}
    \toprule
    \multicolumn{2}{c}{Statistic} & \makecell{SINS/DVL} & \makecell{SINS/DVL\\(Sage-Husa)} & \makecell{SINS/DVL\\(VBAKF)} & \makecell{SINS/DVL\\(ACP)} \\
    \midrule
    \multirow{3}{*}{$\delta P_N$} & Median & -2.0353 & 0.9744 & 0.8070 & 0.3292 \\
    & IQR & 3.7418 & 1.7265 & 1.5492 & 0.7643 \\
    & $P_{95}(|e|)$ & 5.1896 & 2.3315 & 1.9955 & 1.1386 \\
    \multirow{3}{*}{$\delta P_E$} & Median & 0.3261 & -1.0401 & -0.7044 & -0.0068 \\
    & IQR & 1.5626 & 0.8645 & 0.9576 & 0.4607 \\
    & $P_{95}(|e|)$ & 4.1658 & 1.7244 & 1.3958 & 0.7156 \\
    \multirow{3}{*}{$\delta P_D$} & Median & -0.1800 & -0.8701 & -0.8673 & -0.0759 \\
    & IQR & 0.3395 & 0.2045 & 0.1783 & 0.1656 \\
    & $P_{95}(|e|)$ & 0.5263 & 1.1759 & 1.1686 & 0.3980 \\
    \bottomrule
  \end{tabular}
\end{table}

\subsection{Comparison and Analysis}
\label{secComparisonAnalysis}

Unlike the residual-driven Sage-Husa and VBAKF benchmarks, ACP constructs the transformed covariance before the measurement update using the adopted body-frame DVL noise model and predicted attitude-error covariance.
Within ACP, the attitude-related block in the observation matrix represents the first-order effect of attitude error on the projected DVL velocity, while the covariance term in \eqref{eq_cov_with_phi_expect} represents its statistical effect on measurement weighting. Together, these terms keep the projected observation and its covariance consistent when the body-to-navigation-frame transformation depends on uncertain attitude.
For the controlled field data, the lower ACP error metrics are consistent with more effective DVL measurement weighting than that provided by the two residual-driven adaptive-covariance benchmarks under the evaluated conditions.

The results should still be interpreted as a reduction in error growth rather than the removal of dead-reckoning drift.
No absolute underwater position update is introduced during the navigation intervals considered here, and the SINS/DVL solution remains affected by residual inertial-sensor errors, DVL velocity uncertainty, calibration residuals, and environmental disturbances.
ACP does not change the absolute-position observability of the system. Instead, it modifies the stochastic weighting of the DVL velocity update by propagating attitude uncertainty into the navigation-frame measurement covariance.

The evaluation is also bounded by the assumptions of the derivation and the test conditions.
The covariance propagation requires a prescribed body-frame DVL noise covariance and relies on a first-order attitude-error approximation.
Therefore, its direct application is most appropriate after coarse alignment and when the attitude error remains within the small-angle regime.
The present field evaluation does not cover deep-water operation, strong or rapidly varying currents, complex or weakly reflective seabeds, severe or prolonged DVL bottom-lock degradation, severe initial misalignment, or high-dynamic maneuvers.
Such scenarios may require additional mechanisms, such as robust measurement screening, outage handling, or adaptive noise modeling.
Future experiments will evaluate ACP in more diverse underwater environments, cover a wider range of attitude-uncertainty conditions, and incorporate synchronized DVL-quality and environmental records to assess acoustic and hydrodynamic effects.

\section{Conclusion}

\label{secConclusion}

This paper addressed the stochastic inconsistency that arises when a body-frame DVL velocity measurement is projected into the navigation frame using an uncertain SINS-computed attitude.
The proposed method accounts for attitude uncertainty in both the projected velocity observation and its measurement covariance.
The attitude-error-aware observation model represents the effect of attitude error on the projected velocity, while ACP propagates the body-frame DVL covariance at the matrix level and augments it with a closed-form term derived from the predicted attitude-error covariance.

In simulation, compared with SINS/DVL (Sage-Husa), SINS/DVL (ACP) reduced the position RMSE from 14.75 m to 1.01 m and the maximum position error from 20.46 m to 2.38 m.
In the field experiment, compared with SINS/DVL (VBAKF), ACP reduced the northward, eastward, and downward position RMSE values by 48.5\%, 55.0\%, and 74.8\%, respectively.
Under the evaluated simulation and controlled surface-water conditions, these findings support covariance-matrix-level uncertainty propagation for weighting DVL velocity observations whose coordinate projection depends on uncertain attitude information. More broadly, the same propagation structure can be extended to other vector measurements whose coordinate mapping depends on uncertain attitude, providing a model-based route to covariance assignment without changing the underlying filter architecture. Its performance across other vehicle platforms, sensor configurations, broader attitude-uncertainty regimes, and more diverse underwater environments remains to be evaluated.

\bibliographystyle{IEEEtran}
\IfFileExists{IEEEabrv.bib}{
\bibliography{IEEEabrv,refs}
}{
\bibliography{refs}
}

\end{document}